\documentclass[runningheads]{llncs}
\usepackage{graphicx}
\usepackage{cite}
\usepackage{amsmath,amssymb,amsfonts}
\usepackage{algorithmicx}
\usepackage[misc]{ifsym}
\usepackage{textcomp}
\usepackage{xcolor}
\usepackage[ruled]{algorithm2e}
\usepackage{algpseudocode}
\usepackage{diagbox}
\usepackage{multirow}
\usepackage{graphicx}
\usepackage{subfigure}

\begin{document}
\title{Evolving Deep Neural Networks for Collaborative Filtering}
%
%
\author{Yuhan Fang \and Yuqiao Liu\and
Yanan Sun(\Letter)}
\authorrunning{Y. Fang et al.}
%
\institute{College of Computer Science, Sichuan University, Chengdu, China\\
\email{fangyuhan0719@gmail.com;}
\email{lyqguitar@gmail.com;}
\email{ysun@scu.edu.cn}
}
\toctitle{Evolving Deep Neural Networks for Collaborative Filtering}
\tocauthor{Yuhan~Fang, Yuqiao~Liu, Yanan~Sun}

\maketitle 
\begin{abstract}
Collaborative Filtering (CF) is widely used in recommender systems to model user-item interactions. With the great success of Deep Neural Networks (DNNs) in various fields, advanced works recently have proposed several DNN-based models for CF, which have been proven effective. However, the neural networks are all designed manually. As a consequence, it requires the designers to develop expertise in both CF and DNNs, which limits the application of deep learning methods in CF and the accuracy of recommended results. In this paper, we introduce the genetic algorithm into the process of designing DNNs. By means of genetic operations like crossover, mutation, and environmental selection strategy, the architectures and the connection weights initialization of the DNNs can be designed automatically. We conduct extensive experiments on two benchmark datasets. The results demonstrate the proposed algorithm outperforms several manually designed state-of-the-art neural networks.


\end{abstract}
\section{Introduction}
Collaborative Filtering (CF) is a prevalent technique used by recommender systems~\cite{saleem2008collaborative}. The fundamental assumption of CF is that people with similar tastes tend to show similar preferences on items. Therefore, the purpose of CF is to match people with similar interests and make recommendations, by modeling people’s preferences on items based on their past interactions.
Generally, CF algorithms can be divided into two following different categories.

\textbf{1) Matrix Factorization (MF) Algorithms:} MF works by modeling the user’s interaction on items~\cite{koren2009matrix}. Researchers have devoted much effort to enhancing MF and proposed various models such as the Funk MF~\cite{ agarwal2009regression} and the Singular Value Decomposition ++(SVD++)~\cite{cao2015distributed}, \textit{etc.} Funk MF factorizes the user-item rating matrix as the product of two lower dimensional matrices. SVD++ has the capability of predicting item ranking with both explicit and implicit feedback. Unfortunately, the cold start problem, causing by insufficient data, constitutes a limitation for MF algorithm~\cite{bobadilla2012collaborative}.

\textbf{2)Deep Learning Algorithms:} The introduction of deep learning and neural methods into CF tasks has been proven effective by various models, for example, the DeepFM~\cite{guo2017deepfm} and the NeuMF~\cite{he2017neural}. Combining DNN with Factorization Machines (FM), DeepFM has the ability to model both the high-order and low-order feature interactions. In addition, NeuMF generalized the traditional MF with the neural network methods and constructed a nonlinear network architecture. Both these two models can maximize the feature of the finite interaction data and improve the recommendation precision.

Although deep learning methods have shown promising results, designing the network architecture is a challenging and time-consuming task for most inexperienced developers. Therefore, it is worthwhile to extend the automatic method to the network architecture design, ensuring that people who have no domain knowledge of DNNs can benefit from deep learning methods. To this end, we provide an algorithm based on Genetic Algorithm (GA)~\cite{ashlock2006evolutionary}, to automatically design a competitive neural network with efficient weight initialization approaches.

To sum up, the contributions of the proposed Evolve-CF are summarized as follows: \textbf{1) Variable-length encoding strategy:} To represent DNNs in GA, we design a variable-length encoding strategy, which is more suitable to automatically determine the optimal depth of neural networks than traditional fixed-length encoding strategy. \textbf{2) Weights encoding strategy:} To automatically choose the appropriate weights intialization methods, we investigate the weights encoding strategy to optimize the weights efficiently during evolution. \textbf{3) Effective genetic operators:} To simulate the evolution process and increase population diversity, we propose effective genetic operators like crossover and mutation which can cope with the proposed gene encoding strategy. \textbf{4) Improved slack binary tournament selection:} To increase the overall fitness of the whole population during evolution, we improve the slack binary tournament selection to choose promising individuals from the parent population.

\vspace{-0.2cm}
\section{Background}
\vspace{-0.2cm}
\subsection{Skeleton of DNNs}
\label{skeleton_of_dnn}
The skeleton of DNNs for CF consists of the embedding layer, the hidden part, and the output layer. To begin with, the embedding layer is fixed as the first layer to take effect in dimensionality reduction and learn the similarity between words~\cite{Mikolov2013}. Multiple sequential blocks, which are widely used in DNNs for CF, make up the hidden part. Each block contains a full connection layer, a Rectified Linear Unit (ReLU) ~\cite{glorot2011deep} and a dropout layer. A full connection layer is capable of learning an offset and an average rate of correlation between the output and the input. Meanwhile, both the ReLU and the dropout layer can make the model less likely to cause overfitting~\cite{glorot2011deep}.

Besides the arrangement of the blocks, the parameters of each layer affect the performance of DNNs as well, such as the number of neurons in the full connection layer, the dimension of embedding vectors in the embedding layer and the dropout rate in the dropout layer. Both the architecture and the parameters of each layer can be designed automatically.

\subsection{Weights Initialization}
\label{weights_initialization}
In deep learning tasks, weights initialization plays a significant role in model convergence rate and model quality, allowing the loss function to be optimized easily~\cite{Sun2019}. Generally, there are three common categories of initialization methods that are widely used now: \textbf{1) Random Initialization. (\emph{R})} This method uses the probability distribution as the initializer. Nevertheless, without experience and repeated tries, it’s hard for the designers to select the hyper-parameters in the probability model. \textbf{2) Xavier Initialization~\cite{Glorot2010}. (\emph{X})} This method presents a range for uniform sampling based on neuron saturation. While the applicable activation function is limited, which should be a linear function. \textbf{3) Kaiming Initialization~\cite{He2015a}. (\emph{K})} This method proposes an initialization works on ReLU, which is a more prevalent activation function in DNNs. However, its effectiveness highly relies on the architectures of DNNs.

Taking all the advantages and disadvantages into consideration, the designers may have trouble in deciding which weights initialization approaches to choose when designing the neural network for a specific CF task. As a consequence, incorporating searching for weights initialization approaches into GA is needed. Thus the weights initialization approaches can be designed automatically.

\section{The Proposed Algorithm}
Algorithm~\ref{alg_the_proposed_algorithm} displays the framework of the proposed Evolve-CF. Firstly, a population is randomly initialized with the predefined population size, and each individual in the population is randomly generated according to the proposed encoding strategy. Then the evolution begins to work until the generation number exceeds the maximal value defined ahead. In the course of evolution, the fitness of each individual is estimated through a specific method which is selected based on the dataset. After that, by means of the improved slack binary tournament selection method, parent individuals are selected from the population to further conduct the genetic operations consisting of crossover and mutation. Thereafter the offspring is generated. Next, environmental selection takes effect to choose individuals from both the parent population and the generated offspring population, and thereby create the next generation. Then the next round of evolution begins to work. When the whole evolution process is over, the expected best individual is selected, and then we build the DNN decoded from the individual for the final training.

\vspace{-0.3cm}
\begin{algorithm}[h]
\scriptsize
\caption{Framework of the Proposed Algorithm}
\label{alg_the_proposed_algorithm}
\LinesNumbered
\KwIn{The population size, the maximal generation number $s$ }
\KwOut{The best individual}
$P_0$ $\gets$ Randomly initialize the population using the proposed gene encoding strategy\;
\label{alg_framework_line_1}
$t$ $\gets$ $0$\;
\label{alg_framework_line_2}
\While{$t<$s}{
\label{alg_framework_line_3}
Evaluate the fitness of each individual in $P_t$\;
\label{alg_framework_line_4}
$Q_t$ $\gets$ Choose offspring from the selected parent individuals using the proposed genetic operations\;
\label{alg_framework_line_5}
$P_{t+1}$ $\gets$ Environmental selection from $P_t \cup Q_t$\;
\label{alg_framework_line_6}
$t$ $\gets$ $t+1$\;
\label{alg_framework_line_7}
}
\textbf{return} The individual having the best fitness in $P_t$;
\end{algorithm}

\subsection{Gene Encoding Strategy}
\label{gene_encoding_strategy}

Since the genetic operators in GA take effect based on encoding individuals, we present an encoding strategy to encode the DNNs.
The architectures of the neural networks, especially the depths~\cite{Delalleau2011}, play decisive roles in the performance of DNNs and the applicable depths are various.
Because the encoding information contains the depth of the DNNs and the hyper-parameters of each layer, and the depth of the neural network is uncertain, the length of encoding should be variational correspondingly. As a result, we propose a variable-length gene encoding strategy being able to find the optimization automatically, freeing designers from constantly adjusting the length.

\begin{figure}[h]
\centering
\includegraphics[width=0.7\columnwidth]{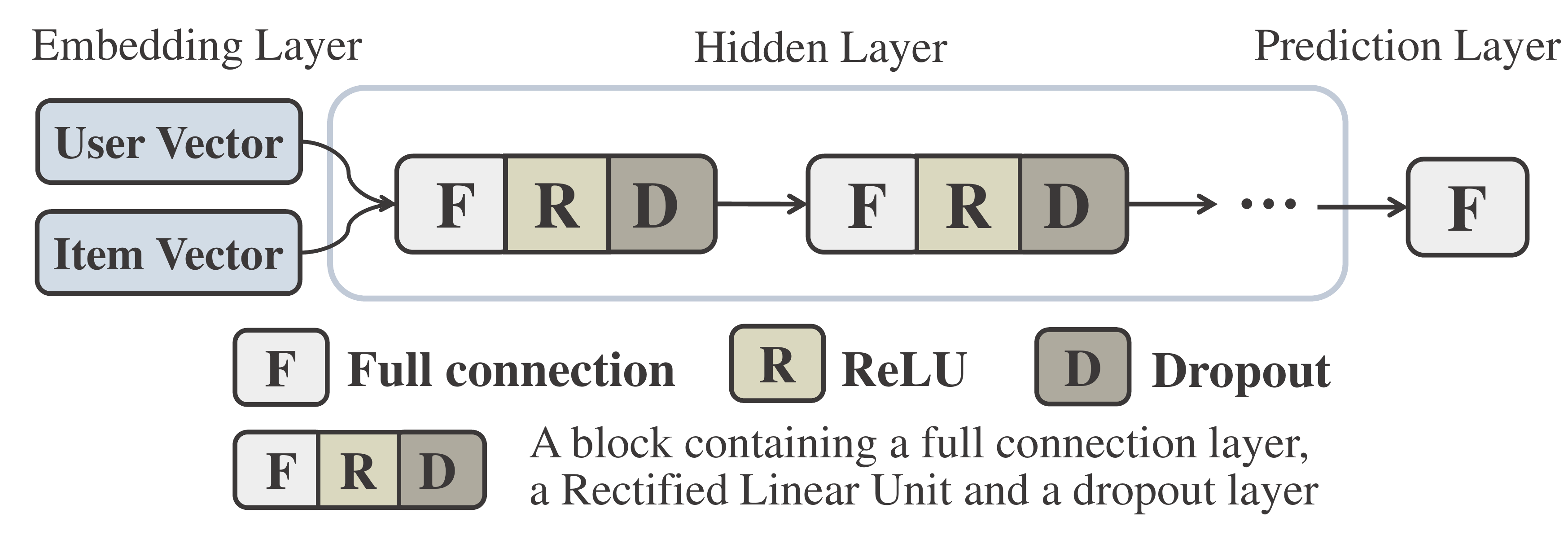}
\caption{An example of the proposed encoding strategy representing a DNN}
\label{example_DNN}
\end{figure}

Via our strategy, the embedding layer and multiple sequential blocks make up the whole DNN. Each block consists of a full connection layer, a ReLU, and a dropout layer. Considering that the last layer determines the prediction results, we set the last block as only a full connection layer. In particular, an example of the proposed strategy representing a DNN is illustrated by Fig.~\ref{example_DNN}.

To efficiently initialize the connection weights, we propose a new method, in which the weights initialization information is explicitly represented as the initialization type of each full connection layer. Simply put, the initialization types are randomly chosen from six types in three categories mentioned in Subsection~\ref{weights_initialization} (\emph{R} with normal distribution $R_n$ and uniform distribution $R_u$, \emph{X} with normal distribution $X_n$ and uniform distribution $X_u$, \emph{K} with normal distribution $K_n$ and uniform distribution $K_u$) and will be optimized during the evolution.

As for the initialization of each individual, we first choose an integer randomly within the range and use it as the length of the current individual. Second, the embedding layer with a predefined input size and a random embedding vector dimension is set as the head of the individual. After that, several blocks are added in sequence to the individual until it meets its predefined length. Finally, the last prediction layer is fixed to make sure the prediction results of each DNN are comparable.
\subsection{Offspring Generation}
\label{offspring_generation}
In GA, the fitness function is usually set based on the task. As for CF, there is no doubt that the fitness function should be able to evaluate the accuracy of recommendations. Because the Normalized Discounted Cumulation Gain (NDCG)~\cite{He2015} can assign higher importance to results at higher ranks on the list, we use it as the fitness evaluation criterion for the task investigated in this work. Specifically, the NDCG is calculated by Equation (\ref{eq1})
\begin{equation}
NDCG@K = \displaystyle\frac{1}{N}\displaystyle[ \displaystyle\frac{1}{D_k}\displaystyle\sum_{i=1}^{K} \frac{2^{r(i)}-1}{\log_2(i+1)}\displaystyle]
\label{eq1}
\end{equation}
where $N$ denotes the total number of involving users, $K$ denotes the length of the top-ranking list, and $r(i)$ refers to the correlation score for the item at position $i$. Respectively, $\sum(\cdot)$ and $D_k$ denote the discounted cumulative gain of the predicted and real ranking list. For NDCG, larger values indicate better performance.

Completing all the fitness evaluations, several individuals are selected as parents in the way mentioned in Subsection~\ref{environmental}. After that, genetic operations like crossover and mutation are performed on the chosen parents. There are two main steps in the mutation operation. The individual is firstly mutated by the Polynomial Mutation (PM)~\cite{Deb2006} in a given probability. Next, the length of the DNN is randomly changed. This change happens in a random position, causing either an increase or decrease in the length of the DNN. As for the proposed crossover operation, we use the Simulated Binary Crossover (SBX)~\cite{Deb1995} for its good performance in local optimization. We first align the heads of two selected individuals and then exchange some of their corresponding blocks according to predefined the crossover rate.

\subsection{Environmental Selection}
\label{environmental}
In the proposed algorithm, we use a developed selection method which is described as Slack Binary Tournament Selection~\cite{Sun2019}. We add an elite mechanism to the traditional binary tournament selection to maintain the diversity and convergence of a population and avoid the premature phenomenon~\cite{Michalewicz2013}. In the course of the environmental selection, firstly, several individuals with the best fitness are directly chosen for the next generation. Secondly, we randomly select two from the rest individuals in the current population using the binary tournament selection, the individuals with larger NDCG are placed into the next population until the population size meets the predefined number. Then, the next iteration starts to work.

\vspace{-0.2cm}
\section{Experiments}

\vspace{-0.3cm}
\subsection{Experimental Settings}
Consistent with previous CF tasks, we use the MovieLens dataset~\cite{harper2015movielens} and Pinterest dataset\footnote{https://sites.google.com/site/xueatalphabeta/academic-projects} as the benchmark datasets.

The settings of the encoded information of each layer in DNN are listed in Table~\ref{encoded_information}. Besides, all the parameter settings of the evolution are specified following the conventions of GA community~\cite{ashlock2006evolutionary}, which are listed in Table~\ref{parameters_GA}. To evaluate the performance of item recommendation, we adopt the \textit{leave-one-out} evaluation~\cite{he2017neural}, which has been widely used in CF models. As mentioned in Section~\ref{offspring_generation}, the NDCG is used as the fitness criterion.

\begin{minipage}{\textwidth}
\begin{minipage}[t]{0.48\textwidth}
\centering
\makeatletter\def\@captype{table}\makeatother
\caption{Encoded information}
\label{encoded_information}
\footnotesize
\begin{tabular}{|c|c|}
\hline
Length of DNN & 4-10 \\ \hline
\begin{tabular}[c]{@{}c@{}}Number of neurons in each \\ full connection layer\end{tabular} & 16-256 \\ \hline
Dropout rate & 0-0.5 \\ \hline
\begin{tabular}[c]{@{}c@{}c@{}}Types of \\ weights initialization \\approaches\end{tabular} &\begin{tabular}[c]{@{}c@{}c@{}}$R_n$, $R_u$\\ $X_n$, $X_u$\\ $K_n$, $K_u$\end{tabular}, \\ \hline
\end{tabular}
\end{minipage}
\begin{minipage}[t]{0.48\textwidth}
\centering
\makeatletter\def\@captype{table}\makeatother
\caption{Parameters of GA}
\label{parameters_GA}
\footnotesize
\begin{tabular}{|c|c|}
\hline
Population size & 16 \\ \hline
Total generation number & 20 \\ \hline
\begin{tabular}[c]{@{}c@{}} Distribution index \\ of SBX and PM \end{tabular} & 1 \\ \hline
Probability of SBX & 0.9 \\ \hline
Probability of PM & 0.2 \\ \hline
Elitism rate & 0.2 \\ \hline
\end{tabular}
\end{minipage}
\end{minipage}

\vspace{0.1cm}


\subsection{Experimental Results}
To show the effectiveness of the proposed algorithm, the following state-of-the-art algorithms are selected as the peer competitors, including deep learning models: NeuMF, MLP~\cite{he2017neural}, and MF models: GMF~\cite{he2017neural}, eALS~\cite{he2016fast}, BPR~\cite{rendle2012bpr}.

The performance of the ranked recommendation list is evaluated by Hit Ratio (HR), which intuitively measures whether the test item is present on the top-k list, and Normalized Discounted Cumulation Gain (NDCG)~\cite{He2015} mentioned in Subsection~\ref{offspring_generation}. For both metrics, larger values indicate better performance.

\begin{figure}[t]
\centering
\subfigure[MovieLens—HR@K]{
\begin{minipage}[h]{0.44\linewidth}
\centering
\includegraphics[width=4.78cm]{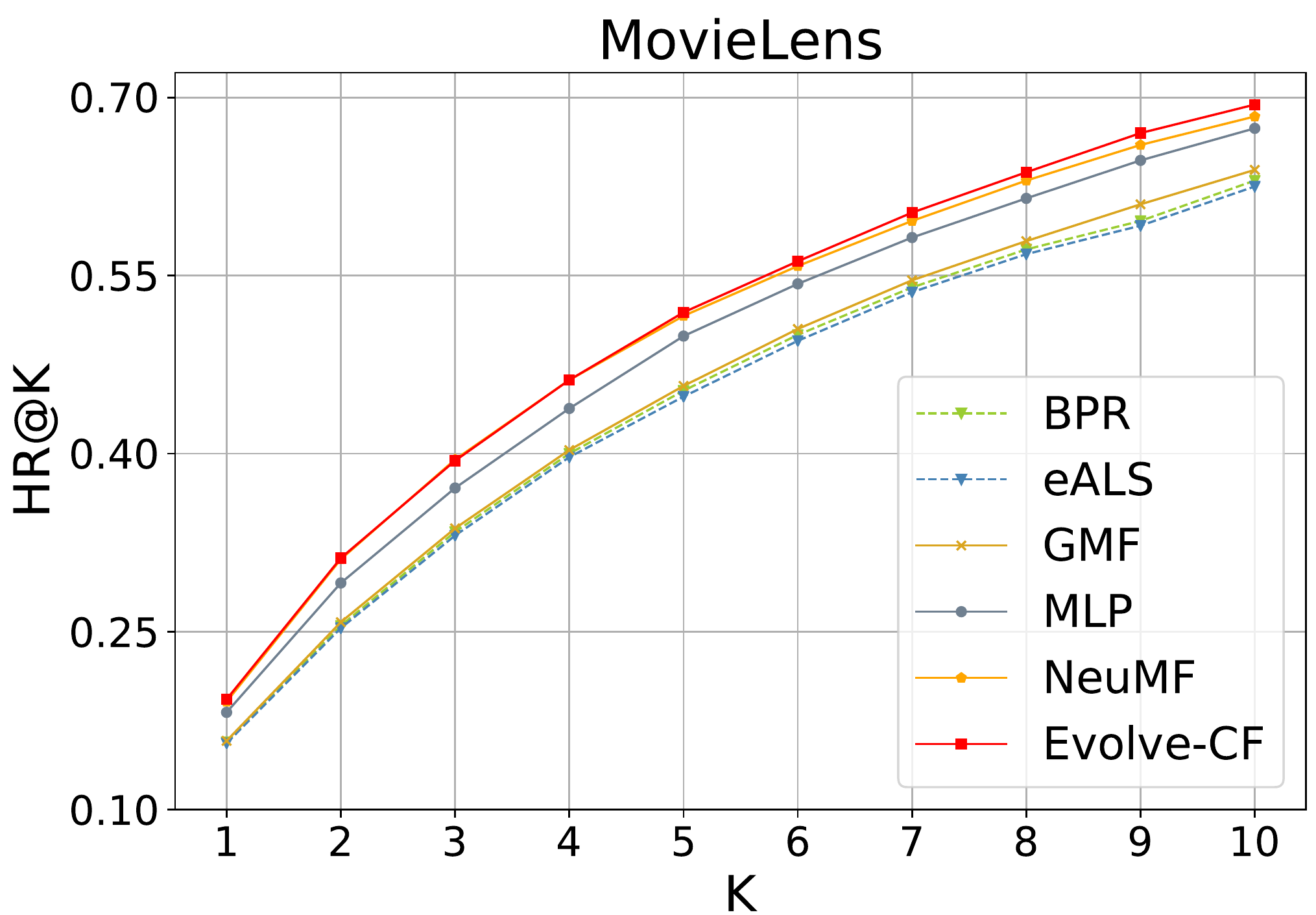}
\end{minipage}%
}%
\subfigure[MovieLens—NDCG@K]{
\begin{minipage}[h]{0.44\linewidth}
\centering
\includegraphics[width=4.78cm]{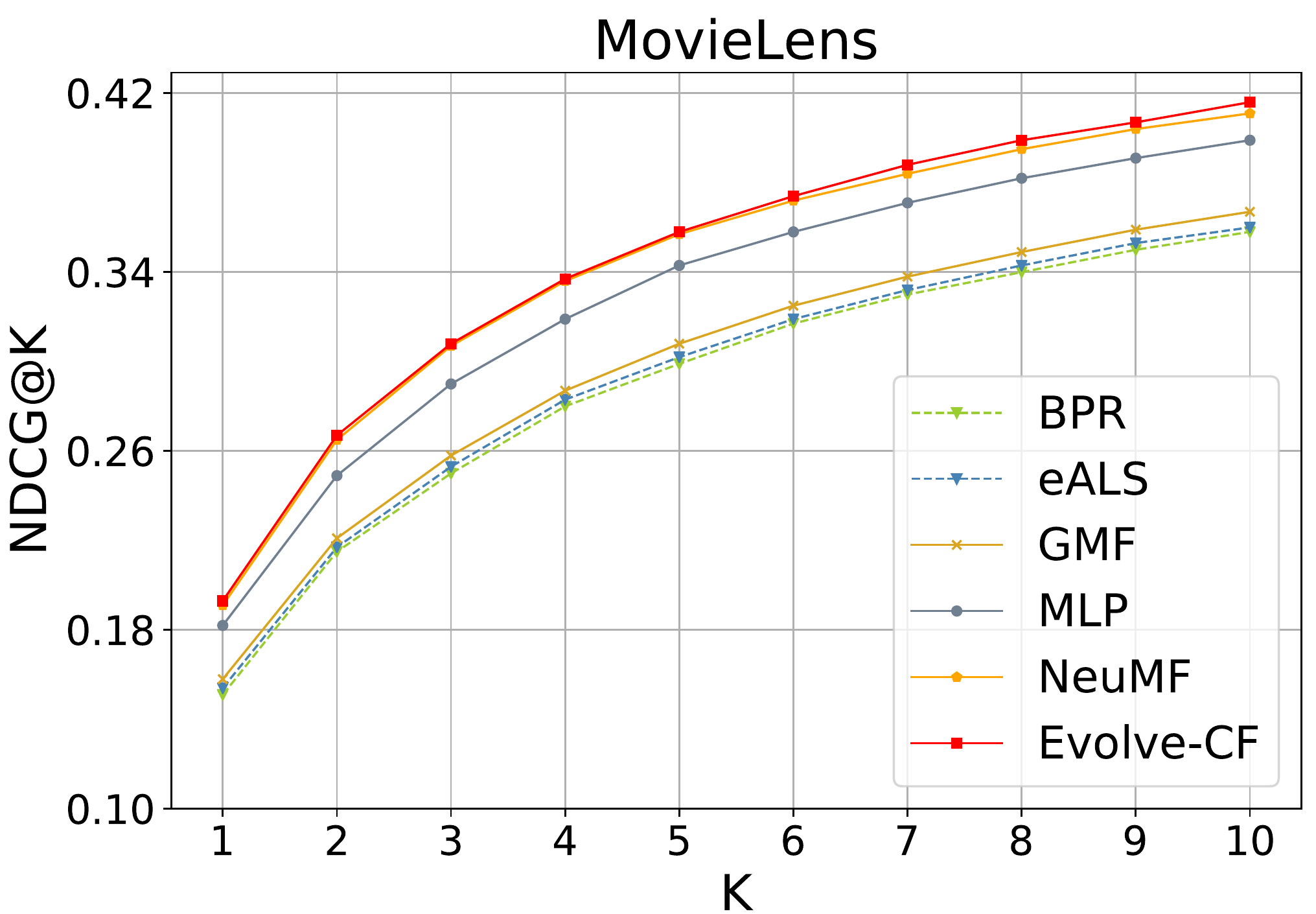}
\end{minipage}%
}%
\vspace{-0.3cm}
\\
\subfigure[Pinterest—HR@K]{
\begin{minipage}[h]{0.44\linewidth}
\centering
\includegraphics[width=4.78cm]{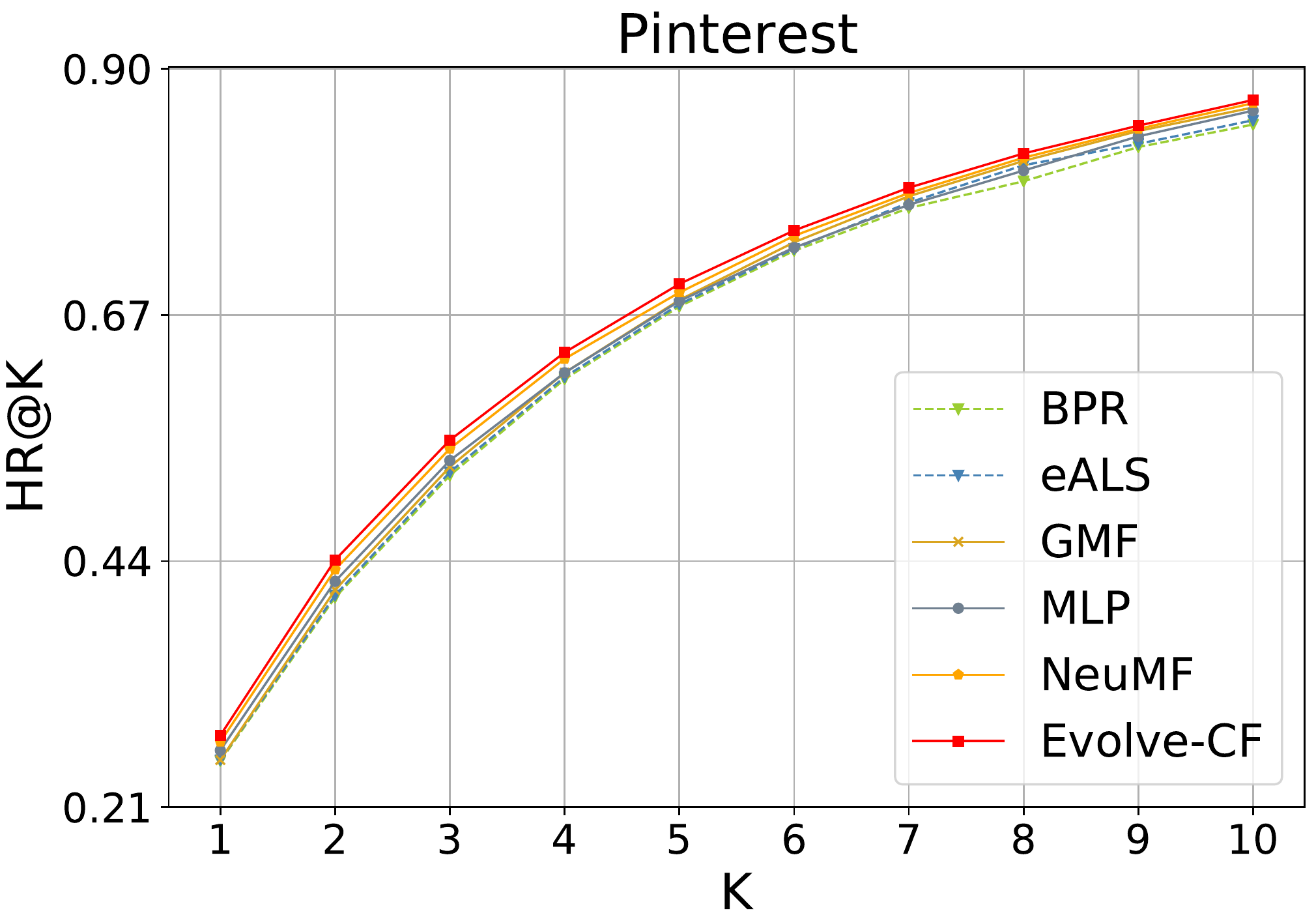}
\end{minipage}%
}%
\subfigure[Pinterest—NDCG@K]{
\begin{minipage}[h]{0.44\linewidth}
\centering
\includegraphics[width=4.78cm]{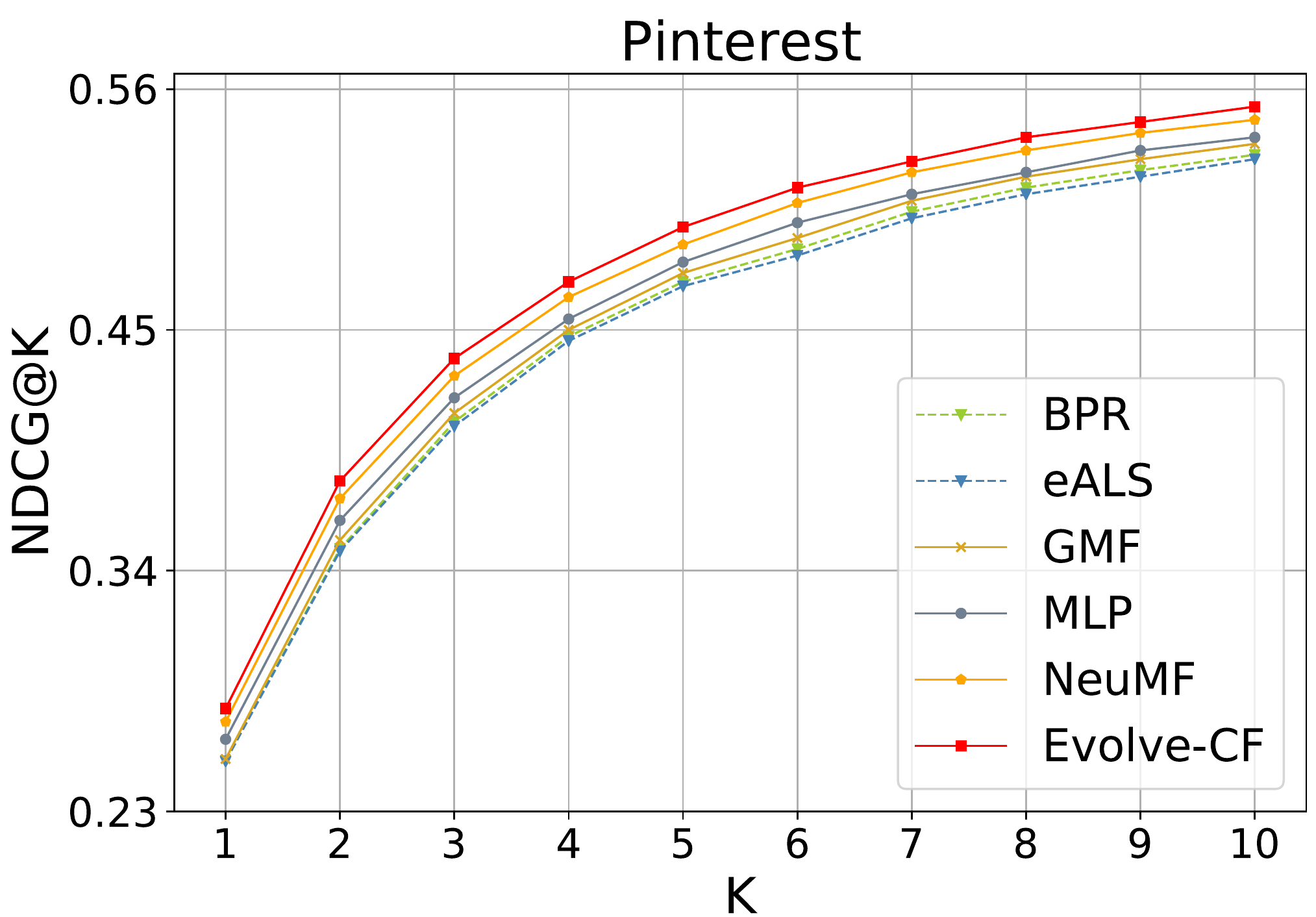}
\end{minipage}%
}%

\centering
\caption{Evaluation of Top-K item recommendation where K ranges from 1 to 10 on the two datasets}
\label{result1}
\end{figure}

Fig.~\ref{result1} shows the performance of Top-K recommended lists with respect to the number of ranking position $K$ ranges from 1 to 10. As can be seen, Evolve-CF demonstrates consistent improvements over other methods across positions.


\begin{table}[h]
\scriptsize
\caption{The best performance of HR@10 and NDCG@10 on the two datasets}
\begin{center}
\begin{tabular}{|p{110pt}<{\centering}|p{33pt}<{\centering}| p{33pt}<{\centering}|p{33pt}<{\centering}|p{33pt}<{\centering}|p{35pt}<{\centering}|p{50pt}<{\centering}|}
\hline
\diagbox{\textbf{Measure}}{\textbf{Method}} & \textbf{BPR} & \textbf{eALS} & \textbf{GMF} & \textbf{MLP} & \textbf{NeuMF} & \textbf{Evolve-CF} \\
\hline
\textbf{Deep Learning Method} &$\times$ &$\times$ &$\times$ & $\surd $ & $\surd $ & $\surd $ \\
\hline
\multicolumn{7}{|c|}{\textbf{MovieLens}} \\ \hline
\textbf{HR@10} & 0.628 & 0.627 & 0.645 & 0.683 & 0.684& \textbf{0.694} \\
\hline
\textbf{NDCG@10} & 0.364 &0.367 & 0.372 &0.408 &0.411 &\textbf{0.416} \\
\hline
\multicolumn{7}{|c|}{\textbf{Pinterest}} \\
\hline
\textbf{HR@10} & 0.856 & 0.861 & 0.864 & 0.862 & 0.868& \textbf{0.871} \\
\hline
\textbf{NDCG@10} & 0.536 &0.522 & 0.541 &0.535 &0.546 &\textbf{0.551} \\
\hline
\end{tabular}
\label{tab1}
\end{center}
\end{table}
\vspace{-0.4cm}
We further focus on the performance of the top 10 recommendation lists. Table~\ref{tab1} shows the best performance of HR@10 and NDCG@10. Obviously, Evolve-CF significantly outperforms both the traditional MF methods and the previous neural network models on two datasets. These experimental results intuitively indicate the effectiveness and generalization of the proposed Evolve-CF.


\vspace{-0.2cm}
\section{Conclusions}
In this work, we explored the evolving strategy to automatically design an appropriate DNN for CF. Our proposed algorithm Evolve-CF provided an improved genetic encoding strategy for encoding the DNN architectures and the weights initialization approaches. Moreover, the genetic operators we selected can effectively find the optimal DNN architecture during the evolution process and the improved binary tournament selection is capable of selecting promising individuals. The results of several experiments on two benchmark datasets demonstrate the effectiveness of the automated DNN over the other CF algorithms.
In the future, further researches will be conducted on finding out whether there are new components that contribute more to the performance of DNNs for CF tasks.

%
%
%
\bibliographystyle{splncs04}

%

\end{document}